\documentclass[runningheads]{llncs}

\usepackage[english]{babel} 
\usepackage[T1]{fontenc} 
\usepackage[utf8]{inputenc} 
\usepackage{graphicx}
\usepackage{color}
\usepackage[algo2e,ruled,vlined]{algorithm2e}
\usepackage{amsmath}
\usepackage{subcaption}

\usepackage{amssymb}
\usepackage{tabularx}
\usepackage{lineno,hyperref}

\usepackage{multirow}
\usepackage{booktabs}

\usepackage{verbatim}

\newcommand{\beq}{\begin{equation}}
\newcommand{\eeq}{\end{equation}}

\newcommand{\ttt}{\boldsymbol{\theta}}

\newcolumntype{L}[1]{>{\raggedright\let\newline\\\arraybackslash\hspace{0pt}}m{#1}}
\newcolumntype{C}[1]{>{\centering\let\newline\\\arraybackslash\hspace{0pt}}m{#1}}
\newcolumntype{R}[1]{>{\raggedleft\let\newline\\\arraybackslash\hspace{0pt}}m{#1}}

\usepackage{xcolor}

\newcommand{\red}[1]{{#1}}

\title{Curriculum semi-supervised segmentation}

\author{Hoel Kervadec, Jose Dolz, \'Eric Granger, Ismail Ben Ayed}

\authorrunning{H. Kervadec et al.}

\institute{\'ETS Montr\'eal}

\begin{document}
    \maketitle

    \begin{abstract}
        This study investigates a curriculum-style strategy for semi-supervised CNN segmentation, which devises a regression network to learn image-level information such as
        the size of the target region. These regressions are used to effectively regularize the segmentation network, constraining the softmax predictions of the unlabeled images
        to match the inferred label distributions. Our framework is based on inequality constraints, which tolerate uncertainties in the inferred knowledge, e.g., regressed region size. It
        can be used for a large variety of region attributes. We evaluated our approach for left ventricle segmentation in magnetic resonance images (MRI), and compared
        it to standard proposal-based semi-supervision strategies. Our method achieves competitive results, leveraging unlabeled data in a more efficient manner and approaching full-supervision performance.

        \red{}
    \end{abstract}

    \section{Introduction}
        \label{sec:intro}

        In the recent years, deep learning architectures, and particularly convolutional neural networks (CNNs), have achieved state-of-the-art performances in a breadth of visual recognition tasks. These architectures currently dominate the literature in medical image segmentation \cite{litjens2017survey}. The generalization capabilities of these networks typically rely on large and annotated datasets, which, in the case of segmentation, consist of precise pixel-level annotations. Obtaining expert annotations in medical images is a costly process that also requires clinical expertise.
        The lack of large annotated datasets has driven research in deep segmentation models that rely on reduced
        supervision for training, such as weakly \red{\cite{lin2016scribblesup,khoreva2017simple,pathak2015constrained,kervadec2019constrained}} or semi-supervised \cite{bai2017semi,sedai2017semi} learning. These strategies assume
        that annotations are limited or coarse, such as image-level tags \red{\cite{papandreou2015weakly,pathak2015constrained}}, scribbles \cite{tang2018regularized} or bounding-boxes \cite{rajchl2017deepcut}.




        In this paper, we focus on semi-supervised learning, a common scenario in medical imaging, where a small set of images are assumed to be fully annotated, but an abundance of unlabeled images is available.
        Recent progress of these techniques in medical image segmentation has been bolstered by deep learning \cite{bai2017semi,baur2017semi,ganaye2018semi,nie2018asdnet,sedai2017semi,zhou2019semi}. Self-training is a common semi-supervised learning strategy, which consists of employing reliable predictions generated by a deep learning architecture to re-train it, thereby augmenting the training set with these predictions
        as pseudo-labels \red{\cite{bai2017semi,pathak2015constrained,rajchl2017deepcut}}.
        Although this approach can leverage unlabeled images, one of its main drawbacks is that early mistakes are propagated back to the network, being re-amplified during training \cite{chapelle2009semi,zhu2009introduction}. Several techniques were proposed to overcome this issue, such as co-training \cite{zhou2019semi} and adversarial learning \red{\cite{dong2018unsupervised,mondal2018few,zhang2017deep}}. Nevertheless, with these approaches, training typically involves several networks, or multiple objective functions, which might hamper the convergence of such models.

        Alternatively, some weakly supervised segmentation approaches have been proposed to constrain the network predictions with global label statistics, for example, in the form of target-region size \cite{Jia2017,kervadec2019constrained,pathak2015constrained}.
        For instance, Jia et \textit{.al} \cite{Jia2017} employed an $\mathcal{L}_2$
        penalty to impose equality constraints on the size of the target regions in the context of histopathology image segmentation. However, their formulation requires the exact knowledge of region size, which limits its applicability. More recently, Kervadec et \textit{al.} \cite{kervadec2019constrained} proposed using inequality constraints, which provide more flexibility, and significantly improves performance compared to cases where learning relies on partial image labels in the form of scribbles. Nevertheless, the values used to bound network predictions in \cite{kervadec2019constrained} are derived from manual annotations, which is a limiting assumption.
        Another closely related work is the curriculum learning strategy proposed in the context of unsupervised domain
        adaptation for urban images in
        \red{\cite{zhang2017curriculum}}. In this case, the authors proposed to match global label distributions over source (\textit{labelled}) and target (\textit{unlabelled}) images by minimizing the KL-divergence between distributions. Finally, it is worth noting that the semi-supervised learning technique in \cite{ganaye2018semi} embeds semantic constraints on the adjacency graph of a given region.

        Inspired by this research, we propose a curriculum-style strategy for deep semi-supervised segmentation, which employs a regression network to predict image-level information such as the size of
        the target region. These regressions are used to effectively regularize the segmentation network, enforcing the predictions for the unlabeled images to match the inferred label distributions. Contrary to
        \red{\cite{zhang2017curriculum}}, our framework uses inequality constraints, which provides greater flexibility, allowing uncertainty in the inferred knowledge, e.g., regressed region size. Another important difference is that the proposed framework can be used for a large variety of region attributes (e.g., shape moments). We evaluated our approach in the task of left ventricle segmentation in magnetic resonance images (MRI), and compared it to standard proposal-based semi-supervision strategies. Our method achieves very competitive results, leveraging unlabeled data in a more efficient manner and approaching full-supervision performance.
        We made our code publicly available\footnote{\url{https://github.com/LIVIAETS/semi_curriculum}}.

    \section{Self-training for semi-supervised segmentation}
        Let $X:\Omega \subset \mathbb{R}^{2,3} \rightarrow \mathbb{R}$ denotes a training image, with $\Omega$ its spatial domain.
        Consider a semi-supervised scenario with two subsets: $\mathcal S = \{(X_i, Y_i)\}_{i = 1, \dots, n}$ which contains a set of images $X_i$ and their corresponding pixel-wise
        ground-truth labels $Y_i$, and $\mathcal U = \{X_j\}_{j = 1, \dots, m}$ a set of unlabeled images, with $m \gg n$. In the fully supervised setting, training is formulated as minimizing the following loss with respect to network parameters $\ttt$:
        \begin{equation}
            \label{eq:fullsup}
            {\cal L}_{Y}(\ttt) = -\sum_{i \in \mathcal{S}} \sum_{p \in \Omega}Y_{i, p} \log S(X_i|\ttt)_p
        \end{equation}
        where $S(X_i|\ttt)_p$ represents a vector of softmax probabilities generated by the CNN at each pixel $p$ and image $i$.
        To simplify the presentation, we consider the two-region segmentation scenario (i.e., two classes), with ground-truth binary labels $Y_{i, p}$ taking values in $\{0, 1\}$, $1$
        indicating the target region (foreground) and $0$ indicating the background. However, our formulation can be easily extended to the multi-region case.
        Common approaches for semi-supervised segmentation \cite{bai2017semi,papandreou2015weakly} generate fake full masks (segmentation proposals) $\tilde{Y}$ for the unlabeled images,
        which are then used
        iteratively for network training by adding a standard cross-entropy loss of the form in Eq. \eqref{eq:fullsup}: $\min_{\ttt} \mathcal{L}_{Y}(\ttt) + \mathcal{L}_{\tilde{Y}}(\ttt)$.
        The process consists of alternating segmentation-proposal generation and updating network parameters using both labeled data and the new generated masks.
        Typically such proposals are refined with additional priors suh as dense CRF \cite{tang2018regularized}.
        However, errors in such proposals may mislead training as the cross-entropy loss is minimized over mislabled points and, reinforcing early mistakes during training, as is well-known in
        the semi-supervised learning literature \cite{chapelle2009semi,zhu2009introduction}.

    \section{Curriculum semi-supervised learning}
\begin{figure}[h!]
            \centering
            \includegraphics[width=.9\textwidth]{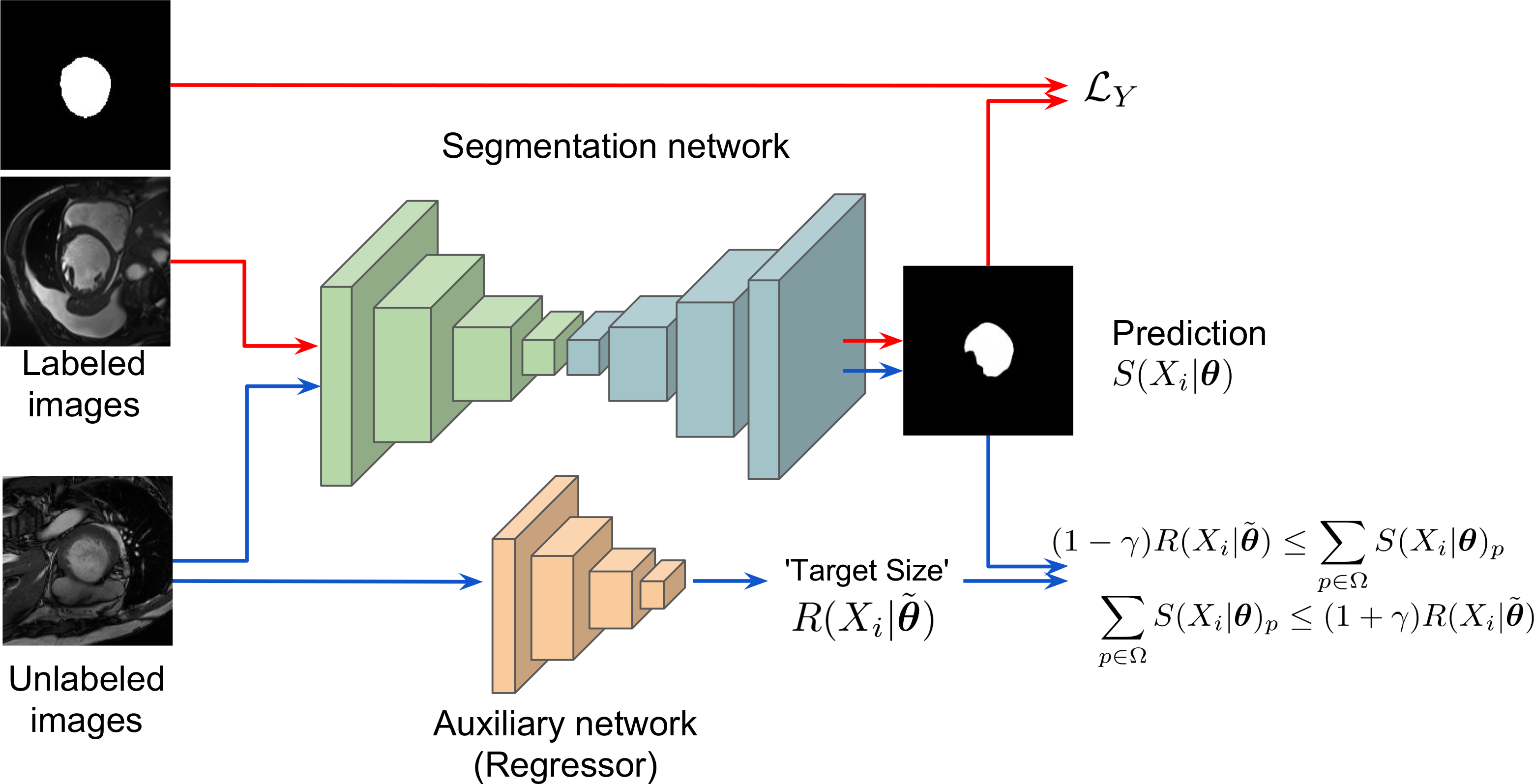}
            \caption{Illustration of our curriculum semi-supervised segmentation strategy.}
            \label{fig:curriculum}
        \end{figure}

        The general principle of curriculum learning consists of solving easy tasks first in order to infer some necessary properties about the unlabeled images. In particular, the first
        task is to learn image-level properties, e.g. the size of the target region, which is easier than learning pixelwise segmentations within an exponentially large label space.
        Then, we use such image-level properties to facilitate segmentation via constrained CNNs. Fig. \ref{fig:curriculum} depicts an illustration of our curriculum
        semi-supervised segmentation. We first use an auxiliary network that predicts the target-region size for a given image. Particularly, we train a regression
        network $R$ (with parameters $\tilde{\ttt}$) by solving the following minimization problem:
        \begin{equation}
            \min_{\tilde{\ttt}} \sum_{i \in \mathcal{S}} \left(R(X_i | \tilde{\ttt}) - \sum_{p \in \Omega} Y_{i,p} \right)^2 .
        \end{equation}
        This amounts to minimizing the squared difference between the predicted size and the actual region size.

        Now we can define our constrained-CNN segmentation problem using auxiliary size predictions $R(X_i | \tilde{\ttt})$:
        \begin{eqnarray}
         \label{eq:constrained-problem}
            && \min_{\ttt} \mathcal{L}_{Y}(\ttt) \nonumber \\
            \text{s.t.} &~~& \forall i \in \mathcal{U}: (1-\gamma) R(X_i | \tilde{\ttt}) \leq \sum_{p \in \Omega} S(X_{i} | \ttt)_p \leq (1+\gamma) R(X_i | \tilde{\ttt}),
        \end{eqnarray}
        where the inequality constraints impose the learned image-level information (i.e., region size) on  the outputs of the segmentation network for unlabeled images, and $\gamma$
        is a hyper-parameter controlling \red{constraints} tightness. We use a penalty-based approach \cite{kervadec2019constrained} for handling the inequality constraints, which accommodates standard stochastic gradient descent. This amounts to replacing the constraints in \eqref{eq:constrained-problem} with the following penalty over unlabeled samples:
        \begin{align}
            \label{eq:loss_unannoated}
            \mathcal{L}_\mathcal{U} (\ttt) &= \sum_{i \in \mathcal{U}} \mathcal{C} \left ( \sum_{p \in \Omega} S(X_{i} | \ttt)_p \right ) \\
            \mathcal{C}(t) &= \begin{cases}
                                (t - (1-\gamma) R(X_i | \tilde{\ttt}))^2 & \text{if } t \leq (1-\gamma) R(X_i | \tilde{\ttt})\\
                                (t - (1+\gamma) R(X_i | \tilde{\ttt}))^2 & \text{if } t \geq (1+\gamma) R(X_i | \tilde{\ttt}) \\
                                0 & \text{otherwise}\\
                           \end{cases}
        \end{align}
        This gives our final unconstrained optimization problem: $\min_{\ttt} \mathcal{L}_{Y}(\ttt) + \lambda \mathcal{L}_\mathcal{U} (\ttt)$, with $\lambda$ a hyper-parameter controlling the
        relative contribution of each term.

    \section{Experiments}
        \label{sec:methods}

        \subsection{Setup}
            \paragraph{\textbf{Data}.}
            \label{sec:data}
            Our experiments focused on left ventricular endocardium segmentation. We used the training set from the publicly available data of the 2017 ACDC Challenge \cite{bernard2018deep}. This set consists of 100 cine magnetic resonance (MR) exams covering well defined pathologies: dilated cardiomyopathy, hypertrophic cardiomyopathy, myocardial infarction with altered left ventricular ejection fraction and abnormal right ventricle. It also included normal subjects. Each exam only contains acquisitions at the diastolic and systolic phases.
            \red{We sliced and resized the exams into $256 \times 256$ images. No additional pre-processing was performed.}

            \paragraph{\textbf{Training}.}
            For the experiments, we employed 75 exams for training and the remaining 25 for validation. From the training set, we consider that \textit{n} images are fully annotated and the pixel-wise annotations of the remaining 75-\textit{n} images are unknown. The \textit{n} images, and their corresponding ground truth, are employed to train both the auxiliary size predictor and the main segmentation network, in a separate way. To validate both networks, we split the validation set into two smaller subsets of 5 and 20 exams, respectively. The training set undergoes data augmentation only to train the size regressor, by flipping, mirroring and rotating (up to 45$^{\circ}$) the original images, obtaining a training set that is 10 times larger.




            \paragraph{\textbf{Implementation details}.}
            \label{sec:implementation}
            We employed ResNeXt 101 \cite{xie2017aggregated} as the backbone architecture for our regressor model, with the squared $\mathcal{L}_2$ norm as the objective function. We trained via standard stochastic gradient descent, with a learning rate of $5\times10^{-6}$, a momentum of $0.9$ and a weight decay of $10^{-4}$,  for 200 epochs. The learning rate was halved at epochs 100 and 150. We used a batch size of 10. We used ENet \cite{paszke2016enet} as the segmentation network, trained with Adam \cite{kingma2014adam}, a learning rate of $5\times10^{-4}$, $\beta_1=0.9$ and $\beta_2=0.99$ for 100 epochs. The learning rate was halved if validation DSC did not improve for 20 epochs. We used a batch size of 1, and $\gamma$ from Eq. \eqref{eq:loss_unannoated} is set at $\gamma=0.1$.
            \red{We did not use any form of post-processing on the network output.}


            \paragraph{\textbf{Comparative Methods}.}
            We compare the performance of the proposed semi-supervised curriculum segmentation approach to several models. First, we train a network using only \textit{n} exams and their corresponding pixel-wise annotations, which is referred to as \textit{FS}. Then, once this model is trained, and following standard proposal-based strategies for semi-supervision, e.g.,
            \red{\cite{bai2017semi}},
            we perform the inference on the remaining 75-\textit{n} exams, and include the CNN predictions in the training set, which serve as pseudo-labels for the non-annotated images (referred to as \textit{Proposals}).
            In this particular case, the training reduces to minimizing the cross-entropy over all the pixels in the manually annotated images and over the pixels predicted as \red{left-ventricle} in the pseudo-labels. Since we investigate how to leverage unlabeled data only by learning from the subset of labeled data, we do not integrate any additional cues during training, such as Conditional Random Fields (CRF)\footnote{Note that the proposal-based methods in \cite{bai2017semi} use CRF to boost performance.}. Finally, we train a model with the exact size derived from the ground truth for each image, as in \cite{kervadec2019constrained}, which will serve as an upper bound, referred to as \textit{Oracle}.

            \paragraph{\textbf{Evaluation}.}
            \label{sec:metrics}
            We resort to the common dice (DSC) overlap metric between the ground truth and the CNN segmentation to evaluate the performances of the segmentation models. More specifically, we report the mean and standard deviation of the validation DSC over the last 50 epochs of training.

        \subsection{Results}
            \label{sec:results}
\begin{figure}
                \centering
                \begin{minipage}{0.49\textwidth}
                    \includegraphics[width=\textwidth]{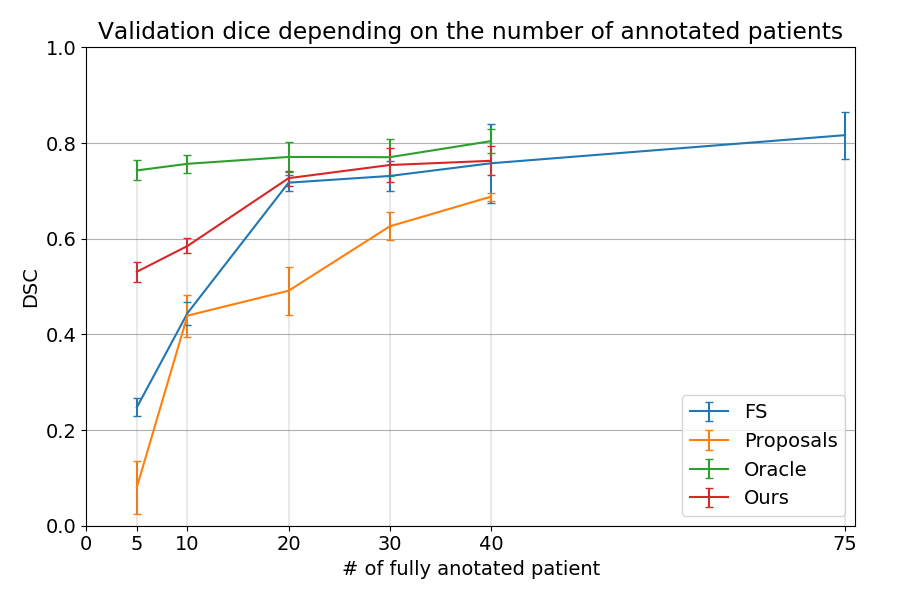}
                    \caption{Mean DSC per method and for several $n$ annotated patients.}
                    \label{fig:abla_patient_avg}
                \end{minipage}\hfill
                \begin{minipage}{0.49\textwidth}
                    \includegraphics[width=\textwidth]{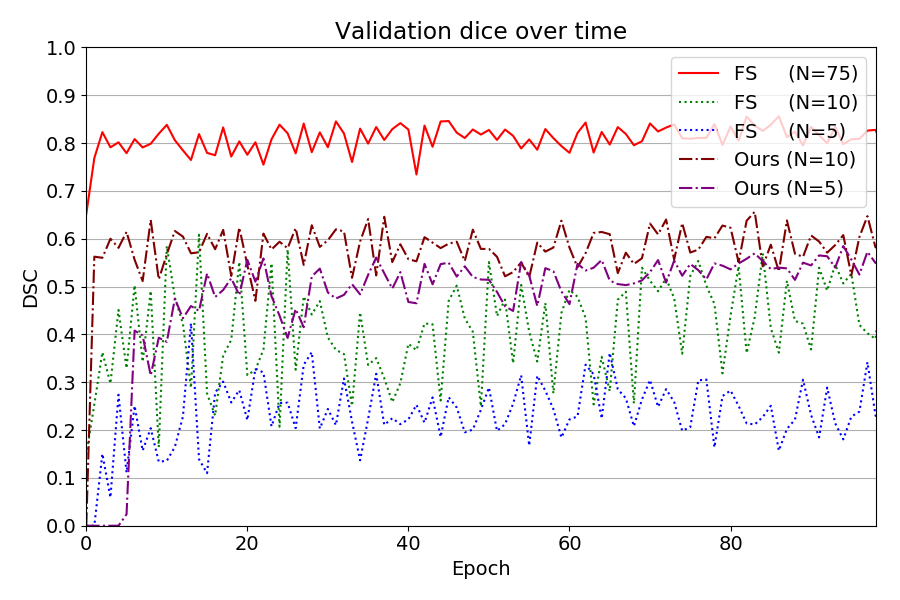}
                    \caption{Validation DSC over time, with a subset of the evaluated models.}
                    \label{fig:val_dsc}
                \end{minipage}
            \end{figure}

            We report in Table \ref{tab:dsc_res} and Fig. \ref{fig:abla_patient_avg} the quantitative evaluation of the different segmentation models. First, we can observe that integrating the size predicted on unlabeled images by the auxiliary network improves the performance compared to solely training from labeled images. The gap is particularly significant when few annotated images are available, ranging from nearly 15 to 25\% of difference in terms of DSC. As more labeled images are available, the proposed strategy still improves the performance of the fully supervised counterpart, but by a smaller margin, which goes from 1 to 3\%. Compared to the \textit{Oracle}, our method achieves comparable results as the number of training samples increases. This suggests that, when few annotated patients are available, having a better estimation of the size helps to better regularize the network. It is noteworthy to mention that in the \textit{Oracle}, the exact size is known for each image, which results in extra supervision compared to the proposed method.
            The \textit{proposals} method achieves the same or worse results than its \textit{FS} counterpart, for all the $n$ values evaluated. These results indicate that $n$ patients are not sufficient to train an auxiliary network that generates usable pseudo-labels, due to the difficulty of the segmentation task. This confirms that training a network on an easier task, e.g., learning the size of the target region, can guide the training in a semi-supervised setting.



           \begin{table}[h!]
                \small
                \centering
                \begin{tabular}{c|c|c|c|c}
                \toprule
                     & \multicolumn{4}{c}{\textbf{Method}} \\
                     \midrule
                    \# labeled patients & \textbf{FS} & \textbf{Proposals} & \textbf{Proposed} & \textbf{Oracle \cite{kervadec2019constrained}} \\
                    \toprule
                    5 & 24.8 (4.9) & 8.1 (0.8) & 53.1 (3.0) & 74.3 (2.5) \\
                    10 & 44.4 (8.3) & 43.9 (2.9) & 58.5 (3.6) & 75.7 (3.9) \\
                    20 & 71.7 (3.2) & 49.1 (5.0) & 72.7 (1.6) & 79.0 (2.5)\\
                    30 & 73.1 (1.7) & 62.6 (4.4) & 75.4 (1.6) & 77.0 (1.9) \\
                    40 & 75.8 (2.4) & 68.8 (5.6) & 76.3 (2.1) & 80.4 (2.1) \\
                    \midrule
                    75 & 81.6 (1.9) & NA & NA & NA \\
                    \bottomrule
                \end{tabular}
                \caption{Quantitative results for the different models. Values represent the mean Dice (and standard deviation) over the last 50 epochs.}
                \label{tab:dsc_res}
            \end{table}

            Evolution of DSC on the validation set over training for some models is depicted in Fig. \ref{fig:val_dsc}.
            From these plots, we can observe that the auxiliary network facilitates the training of a harder task, consistently achieving higher performance and better stability than its \textit{FS} counterpart, especially when few labeled images are available. Regarding the instability of the \textit{FS} method, it may be caused by the small number of samples employed for training, with no other source of information that regularizes the network.


\begin{figure}[h!!]
                \centering
                \includegraphics[width=\textwidth]{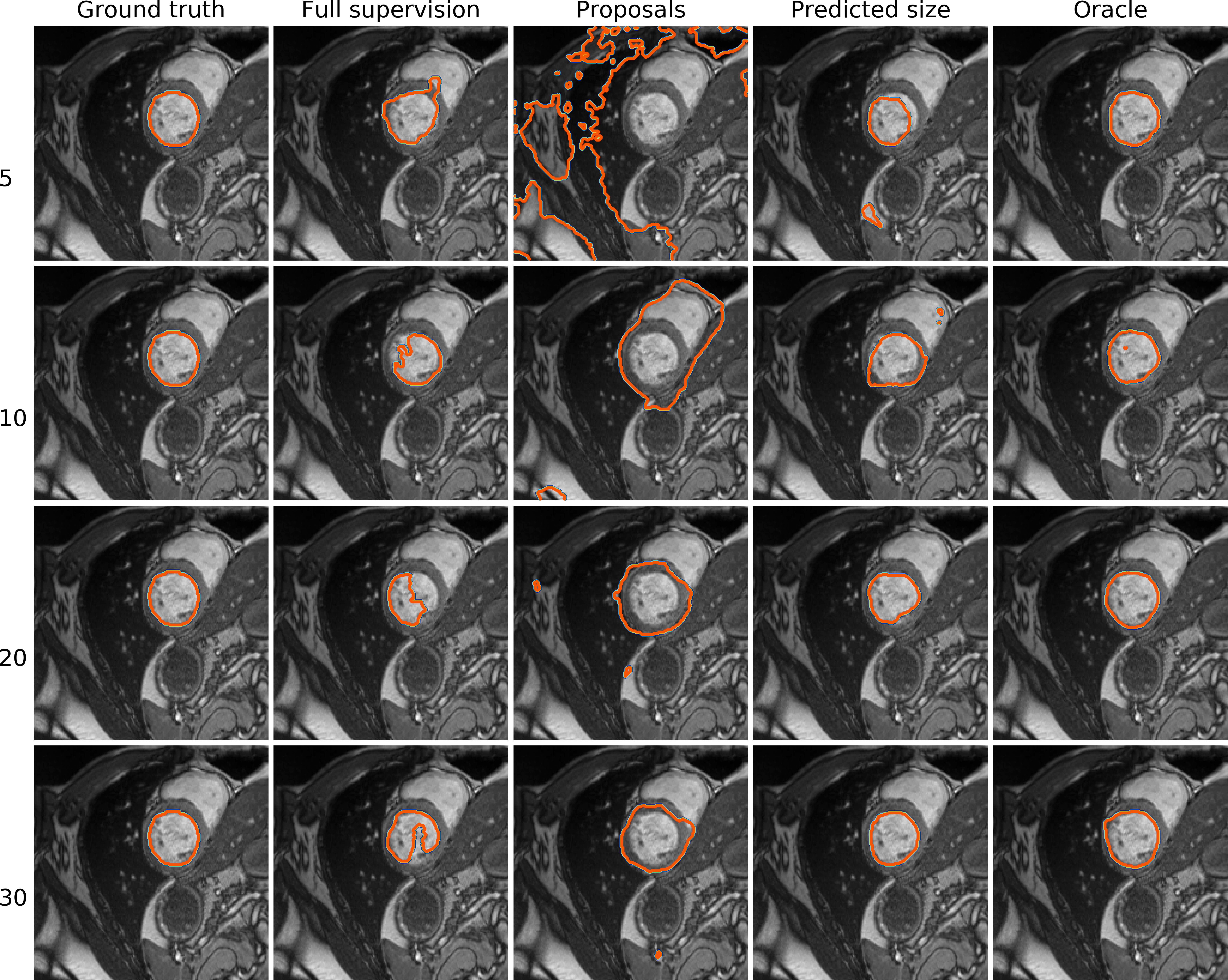}
                \caption{Visual comparison for the different methods, with varying number of fully annotated patients used for training. Best viewed in colors}
                \label{fig:visual_comparison}
            \end{figure}

            Qualitative results are depicted in Fig. \ref{fig:visual_comparison}. Particularly, we show the prediction on the same slice with the different methods and for increasing $n$.
            We first observe that predictions of the \textit{FS} model are very unstable, not clearly improving as more labeled images are included in the training, which aligns with the results found in Fig. \ref{fig:val_dsc}. Then, the \textit{Proposals} approach fails to generate visually acceptable segmentations, even with 30 pixel-wise labeled patients. Although its performance improves with the number of labeled patients used in training, its results are not visually satisfying for any value of \textit{n}. Our curriculum semi-supervised segmentation approach achieves decent results from \textit{n}$=$5. It only requires 20 patients to yield comparable segmentations to those of the \textit{Oracle} and the manual ground truth.

    \bibliographystyle{splncs04}
    \bibliography{paper2227}
\end{document}